\documentclass[conference]{IEEEtran}
\IEEEoverridecommandlockouts
\usepackage{cite}
\usepackage{amsmath,amssymb,amsfonts}
\usepackage{algorithmic}
\usepackage{graphicx}
\usepackage{float}
\usepackage{array}
\usepackage{textcomp}
\usepackage{xcolor}
\def\BibTeX{{\rm B\kern-.05em{\sc i\kern-.025em b}\kern-.08em
    T\kern-.1667em\lower.7ex\hbox{E}\kern-.125emX}}
\begin{document}

\title{CoVis: A Collaborative Framework for Fine-grained Graphic Visual Understanding}

\author{\IEEEauthorblockN{1\textsuperscript{st} Xiaoyu Deng*}
\IEEEauthorblockA{\textit{Fordham University} \\
New York, USA \\
xdeng24@fordham.edu\\ *Corresponding author}
\and
\IEEEauthorblockN{2\textsuperscript{nd} Zhengjian Kang}
\IEEEauthorblockA{\textit{New York University}\\
New York, USA \\
zk299@nyu.edu}
\and
\IEEEauthorblockN{3\textsuperscript{rd} Xintao Li}
\IEEEauthorblockA{\textit{Georgia Institute of Technology} \\
Atlanta, GA, USA \\
xli3204@gatech.edu}
\and
\IEEEauthorblockN{4\textsuperscript{th} Yongzhe Zhang}
\IEEEauthorblockA{\textit{California Institute of Technology}\\
Pasadena, CA, USA \\
yongzhe@caltech.edu}
\and
\IEEEauthorblockN{5\textsuperscript{th} Tianmin Guo}
\IEEEauthorblockA{\textit{New York University}\\
New York, USA \\
tg2374@nyu.edu}
}

\maketitle

\begin{abstract}
Graphic visual content helps in promoting information communication and inspiration divergence. However, the interpretation of visual content currently relies mainly on humans' personal knowledge background, thereby affecting the quality and efficiency of information acquisition and understanding. To improve the quality and efficiency of visual information transmission and avoid the limitation of the observer due to the information cocoon, we propose CoVis, a collaborative framework for fine-grained visual understanding. By designing and implementing a cascaded dual-layer segmentation network coupled with a large-language-model (LLM) based content generator, the framework extracts as much knowledge as possible from an image. Then, it generates visual analytics for images, assisting observers in comprehending imagery from a more holistic perspective. Quantitative experiments and qualitative experiments based on 32 human participants indicate that the $CoVis$ has better performance than current methods in feature extraction and can generate more comprehensive and detailed visual descriptions than current general-purpose large models.

\end{abstract}


\section{Introduction}
Graphic visual content, particularly images, can convey richer information than texts. Compared to videos, it efficiently communicates concise, non-spatiotemporal information, thereby enhancing the efficiency of production and work \cite{pettersson1993visual}. However, an individual's understanding of visual content is often limited by complex factors such as personal life experiences and knowledge background, leading to the effect of information silos \cite{connell2007strategic}. These incomplete interpretations of images may impose potential limitations on the quality of visual guidance for observers. Such limitations can restrict thought processes, confidence, and even cause misunderstandings \cite{worth2016studying}. In turn, these factors can affect the quality of visual content communication and pose obstacles to potential creation \cite{chandler2013seven}, design \cite{chandrasegaran2013evolution}, and optimization \cite{keim2010mastering}.

\begin{figure}[htbp]
\centerline{\includegraphics[width=1.0\columnwidth]{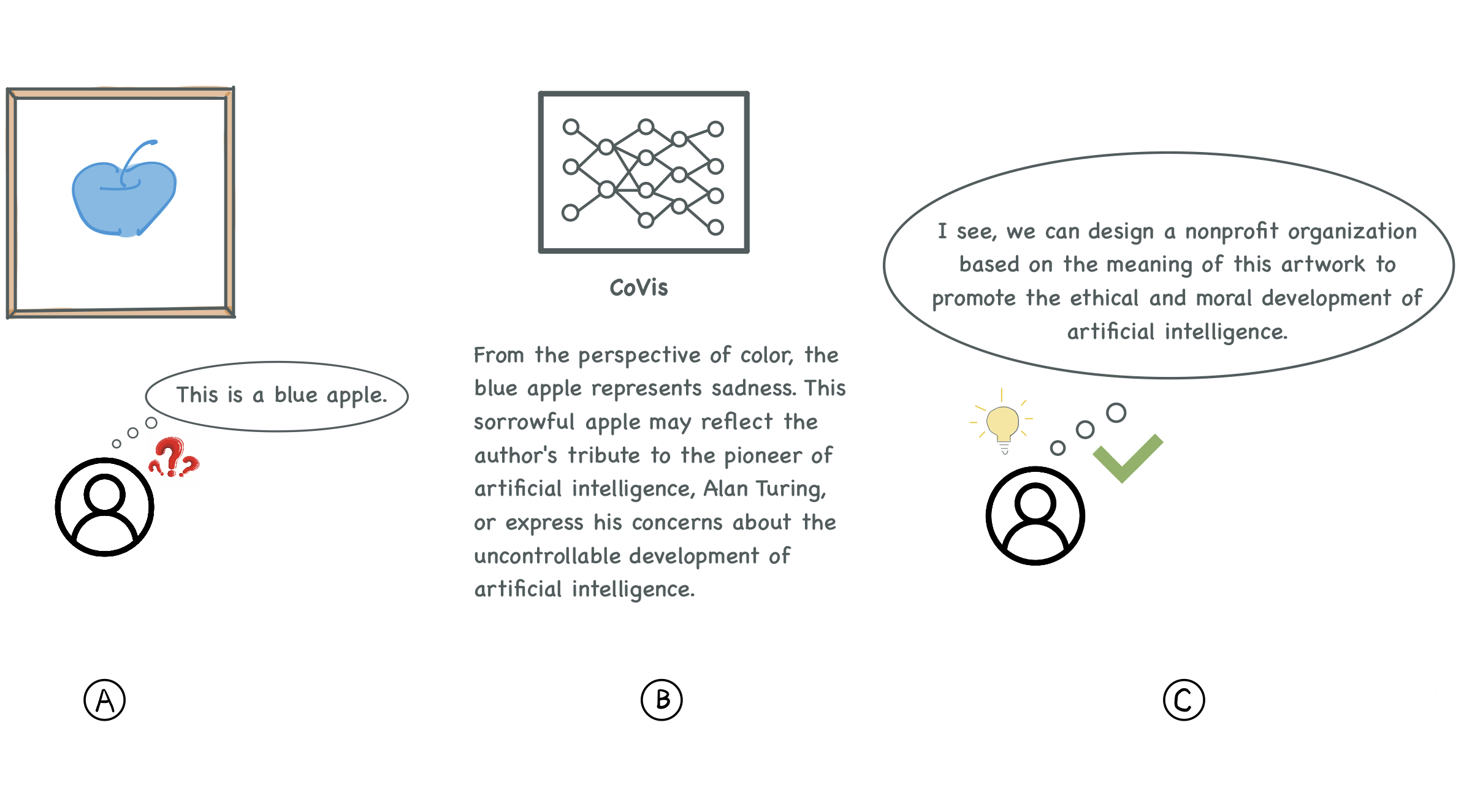}}
\caption{An example of CoVis's inspiration for users.}
\label{fig1}
\end{figure}

Current research in visual image understanding spans a wide range of domains, including image classification \cite{rawat2017deep}, analysis \cite{oberholzer1996methods}, retrieval \cite{datta2008image}, and description \cite{gorokhovatskyi2023search}. However, these methods are often constrained by the data and model scale, limiting them to solving single-dimensional image understanding problems. This limitation stems from the fact that the knowledge base of small-scale models within a singular domain has not yet reached the level of extensiveness observed in human cognition \cite{uchiyama2022cultural}. As a result, these methods struggle to provide comprehensive and objective assistance in visual perception. Consequently, observers' understanding of images largely depends on their own experiential background and cognition, which affects the objectivity and quality of information understanding.

To help break through the limitations of information silos and achieve a more objective and comprehensive understanding of visual semantics, this paper introduces $CoVis$, a collaborative framework for fine-grained visual understanding, to optimizes a cascaded visual segmentation module based on the $FastSam$ (Fast Segment Anything Model)\cite{zhao2023fast} and $U$-$Net$ \cite{huang2020unet} models and bridges it with a multimodal textual content generation model based on $ChatGPT$ $4$. By combining these components with prompt word engineering, the framework generates interpretive text for the main subjects of visual images, assisting observers in understanding images more comprehensively, efficiently, and objectively, as illustrated in Fig. \ref{fig1}. In summary, the main contributions of this paper are as follows:





\begin{itemize}
\item Developed $CoVis$, a collaborative visual understanding framework designed to enhance observers' comprehension of graphic visual content.
\item Implemented dynamic adjustment and optimization of network parameters within the $CoVis$ framework meticulously, ensuring consistently optimal performance and facilitating more accurate and efficient visual analytics.
\item Conducted extensive qualitative and quantitative experiments to verify the effectiveness and usability of the proposed $CoVis$ framework.
\end{itemize}
The paper is structured as follows: Section 2 provides an overview of pertinent research in the field of visual understanding. Section 3 elaborates on the proposed $CoVis$ framework in detail. Section 4 presents the experimental methodologies and outcomes, both quantitative and qualitative. Section 5 encapsulates the research findings and conclusions drawn from this study.

\section{Related Work}

\subsection{Visual Comprehension}

At present, research on visual understanding is extensive, encompassing various domains within computer vision such as image classification, identification, detection, description, and retrieval. For instance, Kim et al. \cite{kim2022transfer} have achieved high-quality medical image classification through a transfer learning-based visual understanding solution and have validated the optimal performance of their proposed method. Gulzar et al. \cite{gulzar2023fruit} addressed the feature recognition challenge in fruit classification with an image recognizer based on the MobileNet V2 network, demonstrating the best performance through extensive experiments. Additionally, Peng et al. \cite{peng2021industrial} proposed an industrial-grade solution framework-based image detector, achieving high-quality fruit ripeness prediction in the agricultural production field, significantly enhancing the efficiency and quality of production work. Li et al. \cite{li2020wavelet} presented a noise-robust image classification framework by integrating a cascaded CNN algorithm, thereby achieving optimal performance. However, these methods are not suitable for comprehensive understanding of image content, as they are primarily aimed at understanding image content in a single domain. They also suffer from the limitations of information silos when assisting visual content observers in understanding images from a more comprehensive perspective. Although some large models have shown high-quality generalization performance in general domains, these models may still have issues such as hallucinations \cite{wang2022generalizing} and randomly generated content \cite{dai2024neural}.

\subsection{Visual Understanding in Human-Machine Collaboration}

Currently, human-computer collaborative systems have significantly enhanced the quality and efficiency of human production, work, and daily life. For instance, Nardo et al. \cite{nardo2020evolution} designed a human-computer collaborative system under the backdrop of Industry 4.0, which has improved production efficiency and reduced resource waste in the industrial production field. Interactive systems \cite{lauretti2023low} based on computer vision for human-computer collaboration can achieve low-cost fruit development quality detection in the agricultural production field, thereby significantly enhancing production efficiency and output. Moreover, the introduction of deep learning technology can make up for the potential risks of errors in manual operation and identification processes. Human-computer collaborative systems \cite{zheng2023intelligent} that integrate CNN and LSTM technologies can accurately recognize building structures in the construction field, thereby verifying the accuracy and rationality of drawings, assisting in correcting potential problems and errors of architects, and thus reducing potential risks and enhancing the reliability of the entire construction project. Furthermore, in the field of education, Othman et al. \cite{othman2010enhancing} proposed a collaborative system for computer science education for college students, which not only ensures interest but also conveys subject knowledge of high quality. In the field of art, Feng et al. \cite{feng2023promptmagician} proposed a collaborative system for creation that achieves high-quality content creation through understanding and analyzing images. In the field of bio-medical research, Sicho et al. \cite{sicho2021genui} developed an interactive collaborative system that can automatically generate potential drug structures, thereby significantly improving the efficiency of drug design and greatly reducing the cycle and consumption of human resources in drug research and development.

\section{Methodology}

\subsection{Overall Design}

\begin{figure*}
\centerline{\includegraphics[width=1.5\columnwidth]{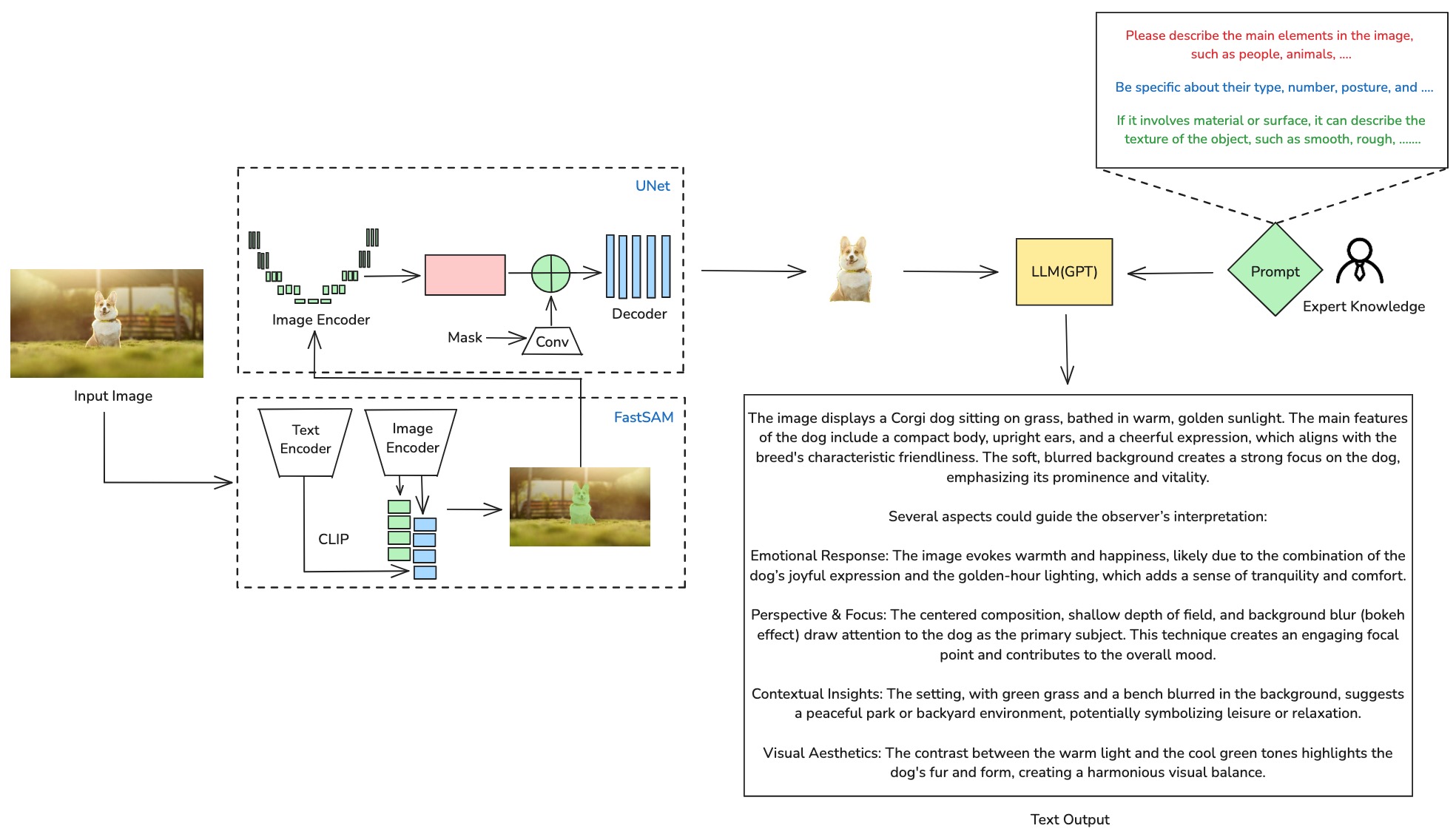}}
\caption{Framework of the proposed CoVis.}
\label{fig2}
\end{figure*}

In this paper, we propose an approach targeting general visual content understanding through a multi-stage image segmentation and language generation method. As illustrated in Fig. \ref{fig2}, the proposed framework includes a coarse-grained and a fine-grained segmentation module, along with a cascaded content generator. Specifically, we employ a $FastSAM$-based module for coarse-grained segmentation. Afterward, a $U$-$Net$ enabled fine-grained image segmentation module is imported and bridge the gap with Large Language Models (LLMs) to produce interpretive text for visual images. To enhance the quality of generated textual content, we incorporate Prompt Engineering techniques. Through cooperation with professional designers, the text of prompt words, including color, composition, connotation, and other dimensions, is deliberately designed, thereby refining the accuracy and consistency of text output.

\subsection{Coarse-grained Segmentation Module}


The backbone network of the coarse-grained segmentation module is $FastSAM$, an advanced pre-trained model optimized for swift image segmentation. It specializes in quickly identifying and segmenting primary objects within an image. $FastSAM$ operates on the principle of feature extraction, generating masks for objects of interest and providing the foundational outline and location necessary for subsequent fine-grained segmentation. A significant advantage of $FastSAM$ is its ability to produce segmentation results without the need for domain-specific data training. This capability allows it to segment images effectively, overcoming the limitations posed by non-gaseous components and high-temperature environments that can affect tools such as laser spectroscopy and CCD cameras.

The $FastSAM$ model leverages the convolutional neural network (CNN) architecture, a cornerstone in deep learning, to extract multi-level feature information from input images through its feature extraction network. The model's architecture consists of the following components:
\begin{itemize}
\item An input layer that receives the original image $I$.
\item A feature extraction module utilizing multiple convolutional layers to distill features.
\item An object recognition module responsible for generating bounding boxes $B$ and masks $M_{\text{group}}$.
\end{itemize}

The feature extraction process of the $FastSAM$ model can be articulated as:


\begin{equation}
F = \phi(I)
\label{eq:1}
\end{equation}

Among them, the extracted feature maps are denoted as \textit{F},  \(\phi\) represents the output of the feature extraction network. Subsequently, the model employs the following formula to generate bounding boxes and masks:

\begin{equation}
B = \gamma(\mathcal{F})
\label{eq:2}
\end{equation}

\begin{equation}
M_{\text{group}} = \delta(F)
\label{eq:3}
\end{equation}

\(\gamma\) and \(\delta\)  represent the boundary box generation function and the mask generation function, respectively. The final coarse-grained segmentation result can be represented as:

\begin{equation}
M_{\text{group}} = \{(B_i, M_i) \mid i \in [1, N]\}
\label{eq:4}
\end{equation}

\subsection{Fine-grained Segmentation Module}


For the fine-grained segmentation module, we have strategically adopted a $U$-$Net$ architecture, which is widely recognized for its proficiency in detailed image segmentation. By synergistically combining $FastSAM$ for coarse segmentation with the $U$-$Net$ for refining the segmentation details, we significantly augment the precision of the overall segmentation process. Specifically, following the acquisition of coarse-grained segmentation, the $U$-$Net$ serves as a fine-grained segmentation module,
further refining the boundaries of the objects. The $U$-$Net$ employs an encoder-decoder architecture, which is adept at progressively restoring the detailed parts of an image, making it particularly suitable for segmentation tasks that demand precise boundaries. By taking the segmentation output from $FastSAM$ as the initial input, $U$-$Net$ is able to accurately refine the object boundaries, yielding high-resolution segmentation outcomes.

The $U$-$Net$ architecture is comprised of an encoder and a decoder. The encoder progressively extracts features, while the decoder utilizes these features to perform high-precision fine-grained segmentation. The input for fine-grained segmentation is the coarse-grained segmentation result $M_{\text{group}}$ generated by $FastSAM$. The output result of the $U$-$Net$ model can be represented as:

\begin{equation}
M_{\text{fine}} = f_{\text{U-Net}}(M_{\text{group}})
\label{eq:5}
\end{equation}

Where \(M_{\text{group}}\) is the fine-grained segmentation result. The fine-grained segmentation process encompasses key technical steps as follows: Firstly, feature extraction is conducted through convolutional layers to capture intrinsic image properties. Subsequently, up-sampling within the decoder progressively restores the spatial information of the image. Finally, the model generates fine-grained masks for each object with precision. Through this sequence of operations, $U$-$Net$ is capable of producing more accurate fine-grained segmentation outcomes based on the initial coarse-grained segmentation.

\subsection{Cascaded Content Generator}

To optimize the generation of interpretive text from visual images, we have integrated a Large Language Model (LLM) enhanced with Prompt Engineering techniques.  This fusion of cutting-edge technologies not only elevates segmentation accuracy but also ensures a highly efficient and structured language output during the visual-to-text transformation process. Specifically, to generate multi-dimensional, fine-grained image descriptions, we propose a 3-step systematic approach for designing prompts based on prompt engineering:

\textbf{Needs Analysis.} In the initial phase, we collaborate with professional designers to delineate the requirements for image description.   This involves identifying key elements within the image, such as color, composition, and connotation, and determining how these elements can be translated into dimensions of textual description.   This step is crucial as it encompasses a deep understanding and analysis of the image content, ensuring that all relevant details and features are captured.

\textbf{Prompt Design.} Based on the outcomes of the needs analysis, we craft a series of prompt words that serve as inputs to the model, guiding it to generate descriptions encompassing the desired dimensions. These prompt words are meticulously constructed to effectively direct the model's attention to the critical features of the image:
\begin{itemize}
\item Color-related: bright, dull, warm tones, cool tones, etc.
\item Composition-related: balanced, symmetrical, dynamic, static, etc.
\item Connotation-related: abstract, realistic, dreamlike, surreal, etc.
\end{itemize}

\textbf{Pilot Experiment Evaluation.} After the initial prompts are designed, we conduct a small-scale experiment to assess their effectiveness. We use these prompts to generate image descriptions and compare the generated descriptions with the actual content of the images.

The segmentation results from $FastSAM$ and $U$-$Net$ are encoded as feature inputs. These features are used to prompt the LLM to generate descriptive text. The process can be formalized as follows:

\begin{equation}
Feature_{Inputs}=Encode(FastSAM, U\_Net)
\end{equation}

\begin{equation}
Descriptive_{Text}=LLM(Feature_{Inputs})
\end{equation}

\begin{equation}
Output=Prompt_{Engineer}(LLM, Feature \_Inputs)
\end{equation}

This method ensures that the generated text is not only accurate but also contextually relevant, providing a comprehensive understanding of the image's content.

\section{Evaluation}
 
\subsection{Experimental Setup}


During the experimental process, we opted for high-performance hardware, including Intel Core i7 processors, NVIDIA GeForce RTX 3080 graphics cards, and the robust Windows 10 Pro version operating system. Code development was efficiently managed through the Jupyter platform. 

\subsection{Quantitative Evaluation on Image Segmentation}

For quantitative analysis, we benchmarked our method against 8 established baselines: PFNet \cite{zhang2020pfnet}, UNet \cite{ronneberger2015u}, SDTC \cite{fan2021rethinking}, MBV3 \cite{howard2019searching}, BASNet \cite{qin2019basnet}, HySM \cite{nirkin2021hyperseg}, U2Net-Tiny \cite{qin2020u2}, and the proposed CoVis. Our evaluation criteria encompassed $F^{max}_{Measure}$, $F_{measure}^{weighted}$, MAE (Mean Absolute Error), $S_{Measure}$, and $E_{Measure}$. 

By comparing our proposed $CoVis$ with these 8 baselines, the results, as shown in Table \ref{tab1}, demonstrate that, our proposed approach has achieved improvements of 1.2\%, 4.7\%, 8.9\%, 1.7\%, and 1.0\% in $F^{max}_{Measure}$, $F_{measure}^{weighted}$, MAE , $S_{Measure}$, and $E_{Measure}$ metrics, respectively, over the state-of-the-art algorithms. This proves that the framework combining $FastSAM$ with $U$-$Net$ possesses robust and high-quality image segmentation performance, surpassing current mainstream advanced methods, laying a good foundation for subsequent image analysis.

\begin{table}[h]
\caption{Comparison evaluation on image segmentation, showing that the proposed CoVis achieves the best performance in all the metrics.}
\centering
\renewcommand{\arraystretch}{1.6}
\begin{tabular}{p{1.5cm}p{1cm}p{1cm}p{0.8cm}p{1cm}p{1cm}}
\hline
\textbf{Methods}&\textbf{$F_{Measure}^{max}$}&\textbf{$F_{Measure}^{weighted}$}& \textbf{MAE} &\textbf{$S_{Measure}$} &\textbf{$E_{Measure}$}\\ 
\hline
U2Net& 0.748& 0.656& 0.09& 0.781&0.823\\

U2Net-Tiny& 0.707& 0.614&   0.095& 0.727&0.012\\
 HySM& 0.734& 0.64& 0.096& 0.773&0.814\\
 BASNet& 0.731& 0.641& 0.094& 0.768&0.816\\
 MBV3& 0.714& 0.641& 0.092& 0.758&0.841\\
 STDC& 0.696& 0.58& 0.103& 0.74&0.817\\
 UNet& 0.692& 0.586& 0.113& 0.745&0.785\\
 PFNet& 0.691& 0.604& 0.106& 0.74&0.811\\
 CoVis& \textbf{0.757}& \textbf{0.687}& \textbf{0.082}& \textbf{0.794}&\textbf{0.831}\\
\hline
\end{tabular}
\label{tab1}
\end{table}


 

\subsection{Ablation Evaluation}
 

In the ablation study, we systematically removed the $SAM$ and $U$-$Net$ detectors from our framework to evaluate their individual contributions, utilizing the same metrics as in our comparative studies.
The studies were designed to meticulously assess the contribution of each integral component of the $CoVis$ framework: the $FastSAM$ and $U$-$Net$ modules. By systematically eliminating them from the complete framework, we can evaluate the performance of the resultant models. The results revealed that the $CoVis$ framework, incorporating both $FastSAM$ and $U$-$Net$, consistently outperformed its counterparts with individual components removed. The results, as presented in the table below, showcasing that our proposed method secured the highest scores across a range of critical metrics, including the  $F^{max}_{Measure}$, MAE, $S_{Measure}$, and $E_{Measure}$.

\begin{table}[h]
\centering
\caption{Image segmentation performance on the ablation evaluation.}

\renewcommand{\arraystretch}{1.6}
\begin{tabular}{p{1.5cm}p{1cm}p{1cm}p{0.8cm}p{1cm}p{1cm}}
\hline
\textbf{Methods}&\textbf{$F_{Measure}^{max}$}&\textbf{$F_{Measure}^{weighted}$}& \textbf{MAE} &\textbf{$S_{Measure}$} &\textbf{$E_{Measure}$}\\ 
\hline
No SAM & 0.692 & 0.586 & 0.113 & 0.745 & 0.785 \\

No U-Net & 0.734 & \textbf{0.696} & 0.097 & 0.760 & 0.802 \\

Ours & \textbf{0.757} & 0.687 & \textbf{0.082} & \textbf{0.794} & \textbf{0.831} \\
\hline
\end{tabular}
\label{tab2}
\end{table}






\subsection{Qualitative Evaluation}

For qualitative evaluation, we engaged $32$ participants from North America and East Asia to rate the content generation of randomly selected images based on satisfaction, accuracy, and creativity.
In order to evaluate the performance of the $CoVis$ in practical applications, we randomly selected $6$ images from the test dataset to evaluate both the advanced Chat $GPT4$-$Mini$ and $Chat$ $GPT4$ models in comparison with our proposed method. Subsequently, we invited $32$ human participants from Asian and American regions to rate descriptions generated by the 3 approaches on satisfaction, accuracy, and creativity, using a 1-to-4 scale (with higher scores indicating better performance). The participants include experts from the art field, designers and randomly recruited users from the Internet. The participants' information and the qualitative assessment results are presented in Tables \ref{HumanID} \& \ref{Human}. The results indicate that our method received the highest scores across all metrics, including satisfaction, accuracy, and creativity.


\begin{table}[h]
\centering
\caption{Information about the participants in qualitative evaluation.}

\renewcommand{\arraystretch}{1.6}
\begin{tabular}{p{2.6cm}p{1.3cm}p{1.3cm}p{1.6cm}}
\hline
\textbf{Category} & Male & Female & $Age_{Average}$\\ 
\hline
Artist                &  4  &  4  &  30.14 \\
Designer              &  5  &  3  &  35.14 \\
Random Participante   &  7  &  9  &  29.93 \\
\hline
\end{tabular}
\label{HumanID}
\end{table}

\begin{table}[h]
\centering
\caption{Participants' ratings of CoVis and other LLM methods on visual description generation.}

\renewcommand{\arraystretch}{1.6}
\begin{tabular}{p{2cm}p{1.6cm}p{1.6cm}p{1.6cm}}
\hline
\textbf{Methods} & Satisfaction & accuracy & creativity\\ 
\hline
GPT4-Mini  &  1.89  &  2.04  &  2.93 \\
GPT4-o     &  2.43  &  2.32  &  2.96 \\
CoVis      &  \textbf{3.32}  &  \textbf{3.25}  &  \textbf{3.39} \\
\hline
\end{tabular}
\label{Human}
\end{table}

\subsection{Generalization Evaluation}

To assess the generalization capabilities of the proposed $CoVis$ method, we conducted experiments across multiple datasets, including $DIS$-$VD$, $ImageNet$-$S$, and $PhenoBench$. As shown in the results table below, our approach exhibits high efficacy and robustness across these diverse datasets. This demonstrates that the $CoVis$ framework can effectively satisfy a wide range of visual content presentation needs, making it a versatile solution for a potentially larger audience.

\begin{table}[h]
\centering
\caption{Generalization evaluation results.}
\renewcommand{\arraystretch}{1.6}
\begin{tabular}{p{1.5cm}p{1cm}p{1cm}p{0.8cm}p{1cm}p{1cm}}
\hline
\textbf{Methods}&\textbf{$F_{Measure}^{max}$}&\textbf{$F_{Measure}^{weighted}$}& \textbf{MAE} &\textbf{$S_{Measure}$} &\textbf{$E_{Measure}$}\\ 
\hline
DIS-VD & 0.757 & 0.687 & 0.082 & 0.794 & 0.831 \\

ImageNet-S & 0.702 & 0.644 & 0.091 & 0.765 & 0.824 \\

PhenoBench & 0.716 & 0.642 & 0.096 & 0.757 & 0.806 \\
\hline
\end{tabular}
\label{tab3}
\end{table}

\section{Conclusion}

This paper introduces the CoVis framework, a collaborative approach for fine-grained graphic visual understanding. The proposed method addresses existing inefficiencies in visual communication by incorporating a cascaded dual-layer segmentation network, complemented by a large-model-based content generator. This integrated framework automates the generation of visual analytics for images, facilitating a more comprehensive understanding of graphic visual contents by extracting a greater amount of information from visual data.

Through extensive quantitative and qualitative experiments, the manuscript demonstrates that the proposed model exhibits enhanced stability, robustness, and insightfulness. These attributes contribute to the improvement of the quality and effectiveness of human-computer interaction within the realm of Computer-Supported Cooperative Work (CSCW) community. Furthermore, this paper reports on the results of generalization experiments, indicating that the proposed framework possesses broad applicability and warrants further exploration. For instance, as a mode of man-machine cooperation for life support, the framework has the potential to significantly assist vulnerable populations, such as the visually impaired, in navigating the visual challenges they encounter in daily life.

However, the current $CoVis$ framework still has some limitations, such as the lack of personalized, targeted content generation strategies. To further improve the efficiency of computer-supported cooperative work, future work would explore incorporating user-specific preferences to enable the targeted generation of stylized visual analysis content.

\bibliographystyle{IEEEtran}
\bibliography{IEEEabrv, references}

\end{document}